\documentclass[conference]{IEEEtran}
\IEEEoverridecommandlockouts

\usepackage{cite}
\usepackage{amsmath,amssymb,amsfonts}
\usepackage{algorithmic}
\usepackage{graphicx}
\usepackage{textcomp}
\usepackage{xcolor}
\usepackage{multirow}
\usepackage{booktabs}
\usepackage{tabularx}
\usepackage{cuted}
\usepackage{enumitem}
\usepackage{subcaption}
\usepackage{authblk}
\usepackage[colorlinks,linkcolor=blue]{hyperref}

\def\BibTeX{{\rm B\kern-.05em{\sc i\kern-.025em b}\kern-.08em
    T\kern-.1667em\lower.7ex\hbox{E}\kern-.125emX}}
\begin{document}

\title{StegOT: Trade-offs in Steganography via Optimal Transport\\
\thanks{\hspace*{-\parindent}\rule{3.5cm}{0.5pt}\\
\IEEEauthorrefmark{1} Correspoding author.
}
}


\author{\textit{Chengde Lin}$^{1}$, \textit{Xuezhu Gong}$^{1}$, \textit{Shuxue Ding}$^{2,1,}$\IEEEauthorrefmark{1}, \textit{Mingzhe Yang}$^{1}$, \textit{Xijun Lu}$^{3}$, \textit{Chengjun Mo}$^{1}$}

\affil{$^{1}$School of Artificial Intelligence, Guilin University of Electronic Technology, Guilin, China}
\affil{$^{2}$College of Information Engineering, Gandong University, Fuzhou, China}
\affil{$^{3}$Medical College, Tianjin University, Tianjin, China}
\affil{$\{$xzgong, mochengjun$\}$@mails.guet.edu.cn $\{$lcd, sding$\}$@guet.edu.cn $\{$ymz, xjunlu$\}$@163.com} 

\date{} 


\maketitle

\begin{abstract}
Image hiding is often referred to as steganography, which aims to hide a secret image in a cover image of the same resolution. Many steganography models are based on generative adversarial networks (GANs) and variational autoencoders (VAEs). However, most existing models suffer from mode collapse. Mode collapse will lead to an information imbalance between the cover and secret images in the stego image and further affect the subsequent extraction. To address these challenges, this paper proposes StegOT, an autoencoder-based steganography model incorporating optimal transport theory. We designed the multiple channel optimal transport (MCOT) module to transform the feature distribution, which exhibits multiple peaks, into a single peak to achieve the trade-off of information. Experiments demonstrate that we not only achieve a trade-off between the cover and secret images but also enhance the quality of both the stego and recovery images. The source code will be released on \href{https://github.com/Rss1124/StegOT}{https://github.com/Rss1124/StegOT}. 
\end{abstract}

\begin{IEEEkeywords}
steganography, optimal transport, information hiding, autoencoders.
\end{IEEEkeywords}

\section{Introduction}

In today's digital age, information security is essential to protect sensitive data. Steganography, as an information-hiding technique, has a core goal of embedding secret data into the carrier without attracting attention. Image steganography is a crucial part of steganography, aiming to embed large-capacity watermark information (such as RGB images) into a carrier while ensuring that the watermark remains invisible. This poses a significant challenge in maintaining a balance between the carrier and the watermark. Many current steganography models based on generative adversarial networks (GANs) and variational autoencoders (VAEs) show promise in practice (e.g.,\cite{song2021enhancing,butora2020steganography,luo2023content}), but mode collapse still poses research challenges.

\begin{figure}[htbp]
\centerline{
\includegraphics[scale=0.3]{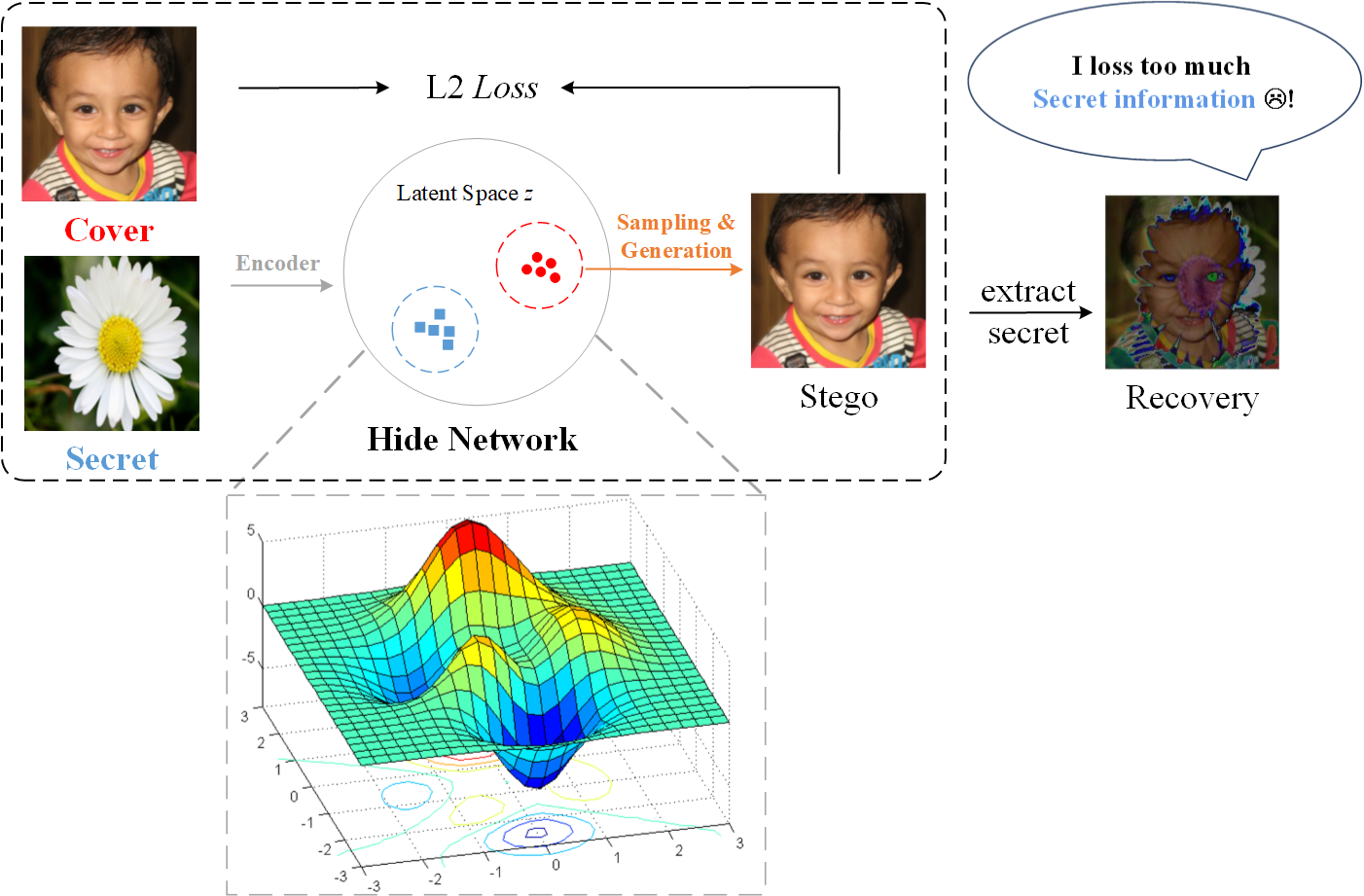}
}
\caption{In GAN-based or VAE-based steganographic models, the decoder is more inclined to sample the information of Cover image, which leads to the imbalance between Cover information and Secret information in Stego images.
}
\label{fig:mode collapse}
\end{figure}

Typically, neural networks use an encoder and a decoder to process data. The encoder maps the input into a latent space, and the decoder extracts latent vectors from it to generate an output. Unfortunately, as highlighted by these works(e.g.,\cite{nagarajan2017gradient,khayatkhoei2018disconnected,xiao2018bourgan}), when the data distribution exhibits multiple modes, the transport mapping of deep neural networks becomes discontinuous. This inherent conflict can lead to mode collapse. As shown in Figure~\ref{fig:mode collapse}, the encoder in the steganography model maps both the cover image and the secret image into the same latent space. The loss function is defined as the difference between the stego image and the cover image, which is typically calculated using a common metric like L2 loss. This design encourages the decoder to preserve the characteristics of the cover image as much as possible when generating the stego image. As a result, the secret image's information is suppressed, leading to difficulties in extracting the watermark from the stego image later on. To solve the trade-off problem caused by mode collapse, this paper introduces an autoencoder-based steganography model integrating optimal transport theory. It uses an autoencoder to map images into latent space and applies discrete optimal transport for discrete mapping, which transforms the multi-peak feature distribution into a single peak. Experimental results demonstrate the effectiveness of the proposed method. The proposed method improves all evaluated indicators and convergence speed. The main contributions of this paper are as follows:
\begin{itemize}[label=\textbullet]
    \item We propose a novel watermarking network based on optimal transport theory, which provides new insight into solving the trade-off between cover and secret information while offering a solid theoretical foundation.
    \item We introduce a new Multi-Channel Optimal Transport (MCOT) model to achieve a better balance between Cover information and Secret information in Stego images, which not only improves the image quality but also improves the ability robustness of Stego images.
    \item We provide a sufficient experimental demonstration that our StegOT outperforms existing state-of-the-art (SOTA).
\end{itemize}

\begin{figure*}[htbp]
\centering
\includegraphics[width=\linewidth]{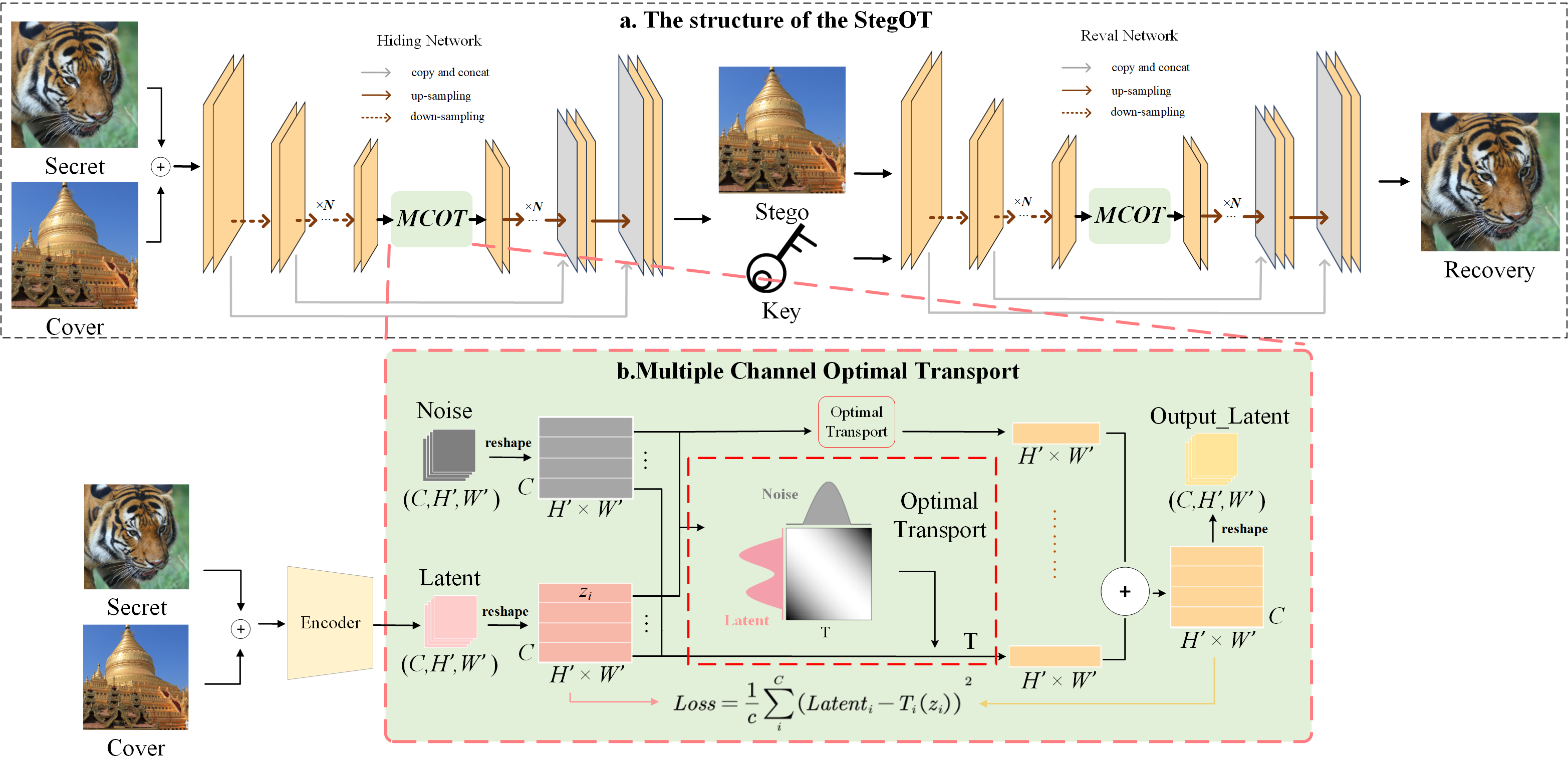}
\caption{The framework of StegOT.   In the hiding network, MCOT transforms a multi-peak feature distribution into a single-peak distribution.   In this way, MCOT achieves a balance between cover information and secret information in the stego image. To protect the watermark image from leaking out, we use the optimal transmission map T generated by MCOT as the key to start the Reval Network.}
\label{fig:mcot}
\end{figure*}

\section{Related Work}
\subsection{Optimal Transport}
Optimal Transport (OT) is a concept in mathematics that originates from the mid-19th century, originally proposed by French mathematician Gaspard Monge. The problem initially focused on determining how to transport a given mass of soil to specified locations at minimal cost. In modern statistics and machine learning, OT is used to determine the optimal transport map between two probability distributions, with one of its most common applications being style transfer (e.g., \cite{Kolkin_2019_CVPR,cheng2023user,peyre2019computational}). However, research on optimal transport in the field of steganography remains highly underexplored.

\subsection{Steganography Model}
In recent years, the rapid evolution of deep learning has driven the development of many steganography techniques. The first use of CNNs for image steganography was introduced by Baluja \cite{baluja2017hiding}. Subsequently, Luo \cite{luo2020distortion} and Volkhonskiy \cite{volkhonskiy2020steganographic} enhanced the perceived quality of stego images by using GANs. Hiding images is a key aspect of steganography. Its purpose is to hide one or more secret images within a carrier image and enable their extraction from the stego image. BalujaNet \cite{baluja2017hiding} has become a significant model in the field, influencing numerous subsequent research works (e.g., \cite{Ahmadi_Norouzi_Karimi_Samavi_Emami_2020,chen2023low,liu2022image,durall2020watch}). Some studies have introduced Invertible Neural Networks (INNs) into reversible tasks such as denoising, scaling, and colorization. Compared to other architectures like autoencoders and GANs, INNs have demonstrated notable advantages. HiNet \cite{jing2021hinet} is the first network to utilize INNs for image hiding by embedding images in the wavelet domain, which greatly improves their performance. Compared to autoencoder-based steganography, INN-based steganography achieves superior results in terms of image quality and data embedding efficiency (e.g., \cite{zhang2024efficient,guo2022invertible}). This is because autoencoder-based steganography often faces challenges in balancing the quality of cover images and secret images.
\par As studied by Brenier \cite{brenier1991polar} and Figalli \cite{figalli2010regularity}, we hypothesize that the imbalance phenomenon is caused by mode collapse. Because the feature distributions exhibit multi-peak characteristics and the loss function design, the decoder in the hiding network tends to extract more information from the cover image while neglecting information from the secret image. This results in mode collapse, disrupting the balance between cover information and secret information in the stego image. To mitigate mode collapse, several recent studies have introduced new objective functions, network architectures, or alternative training schemes (e.g., \cite{zou2023auto, li2021tackling, yu2022hsgan}). However, these approaches often come at the expense of image quality. Inspired by Ferradans \cite{ferradans2014regularized}, we propose an autoencoder-based steganography model that integrates optimal transport theory. Our goal is to transform the multi-peak feature distribution into a single-peak, balancing cover information and secret information.

\section{Proposed Method}
\subsection{Overall Pipeline}
In the feature extraction stage, the encoder maps the cover image and the secret image into latent space after concatenation, then extracts the latent vector. However, the data distribution is highly complex. As a result, the encoder cannot balance the amount of information between the cover image and the secret image. Therefore, we use the discrete optimal transport map to transform the multi-peak distribution into a single-peak distribution, ensuring minimal information loss during the process.
\par
As shown in Figure~\ref{fig:mcot}, our steganography model consists of two key components: i) an autoencoder (AE), with a U-Net structure as the overall framework; and ii) the Multiple Channel Optimal Transport (MCOT) module, which computes the optimal transport map \textit{T} from the source distribution to the target distribution. The inputs to the AE are the cover image $Input_{cover} \in R^{3\times H\times W}$ and the secret image $Input_{secret} \in R^{3\times H\times W}$. These are concatenated across the RGB channels to form a new input $Input_{msg}\in R^{6\times H\times W}$. According to the U-Net design, $Input_{msg}$ will pass through 4 encoder stages, each consisting of a downsampling layer. In the downsampling layer, we double the number of channels and reduce the feature resolution by half. At the bridge stage of U-Net, we use MCOT to map the source distribution to the target distribution. Finally, we reconstruct this feature distribution into an image through 4 decoder stages. To ensure hiding performance, the loss function of our hiding network is defined as follows:\begin{equation}L_h=\sqrt{\lVert I_{cover}-D\left(T\left(E\left(I_{cover}, I_{secret}\right)\right)\right)\rVert ^2}+\epsilon ^2,\label{loss_h}\end{equation}where \textit{E} represents the encoder of the hiding network, \textit{D} represents the decoder of the hiding network, \textit{T} denotes the optimal transport mapping, and \textit{$\epsilon$} is a small positive constant used to prevent the loss from being zero. 
\par 
To ensure recovery performance, the loss function of our revealing network is defined as follows:\begin{equation}L_r=\sqrt{\lVert I_{secret}-D'\left( T'\left( E'\left( I_{stego} \right) \right) \right) \rVert ^2}+\epsilon ^2,\label{loss_r}\end{equation}where \textit{$E'$} represents the encoder of the revealing network, \textit{$D'$} represents the decoder of the revealing network, \textit{$T'$} denotes the optimal transport mapping.

\subsection{Multiple Channel Optimal Transport}
In hiding network, we will initialize a white noise that satisfies a normal distribution $Noise\in R^{C\times H'\times W'}$, and then pass the $Latent\in R^{C\times H'\times W'}$ of the encoder into MCOT together. Here, \textit{C} represents the number of channels in the latent vector, while \textit{$H'$} and \textit{$W'$} represent the height and width of the latent vector, respectively. Then we perform a reshape operation on $Latent$ and $Noise$ to obtain $Latent'\in R^{{C\times N}}$ and $Noise'\in R^{{C\times N}}$, where $N=H'\times W'$. Subsequently, we calculate the discrete mapping $T_i$ from the $Latent'_i$ to the $Noise'_i$, where $i\in\{1,...,C\}$. 
\par
For the characteristic distributions X and Y of $Latent'$ and $Noise'$, our goal is to transfer the distribution of X (multi-peak) to the distribution of Y (single-peak). We define it as an optimal transport problem and the formula is as follows:\begin{equation}\min_{T:X\to Y}\int_Xc(x,T(x))d\mu(x),\label{eq:1}\end{equation}
where $x\subset X\text{ and }y\subset Y$, $c$ is the cost function, which defines the cost per unit mass of transport from position from $x$ to $y$, $\mu(x)$ is a probability measure that describes the distribution of resources in space X.
\par
To solve the transport map \textit{T}, we discretize the continuous optimal transport problem, treating it as a mapping between multiple points. Consequently, we employ a Multi-Layer Perceptron (MLP) to address this linear transformation problem. The linear programming model we constructed is as follows:\begin{equation}\min_{T}\sum_{i,j=1}^Nc(X_i,Y_j)T_{i,j},\label{eq:2}\end{equation} 
where the cost function is defined as $c(X_i,Y_j)=||X_i-Y_j||_2^2$. We use a 2-layer MLP with ReLU activation to try to fit this linear programming problem. To ensure transmission performance, our transport loss is defined as follows:
\begin{equation}L_T=\frac{1}{c}\sum_i^C{\sqrt{\left( Latent_i-T_i\left( z_i \right) \right) ^2}}\label{loss_t}\end{equation} 

\subsection{Loss Function}
The overall loss function of our method is the sum of the hiding loss $L_h$, the reveal loss $L_r$, and the transport loss $L_T$, as follows:
\begin{equation}L_{total}\ =\ L_h+L_r+L_T\label{loss_total}\end{equation} 

\begin{table*}[htbp]
\centering
\caption{benchmark compare of ${Cover}$/${Stego}$ image pair on different datasets}
\label{table:comparison1}
\begin{tabularx}{\textwidth}{lXXXX|XXXX|XXXX}
\toprule
\multirow{2}{*}{} & \multicolumn{4}{c|}{COCO${(256\times256)}$} & \multicolumn{4}{c|}{DIV2K${(256\times256)}$} & \multicolumn{4}{c}{ImageNet${(256\times256)}$} \\
\cmidrule{2-13}
 & PSNR$\uparrow$ & SSIM$\uparrow$ & MAE$\downarrow$ & RMSE$\downarrow$
 & PSNR$\uparrow$ & SSIM$\uparrow$ & MAE$\downarrow$ & RMSE$\downarrow$ 
 & PSNR$\uparrow$ & SSIM$\uparrow$ & MAE$\downarrow$ & RMSE$\downarrow$ \\
\midrule
HiDDeN\cite{zhu2018hidden} & $33.33$ & $0.9109$ & $7.97$ & $10.98$ & $34.27$ & $0.9216$ & $7.67$ & $10.31$ & $33.28$ & $0.9064$ & $8.29$ & $11.30$ \\
Weng\cite{weng2019high} & $36.31$ & $0.9161$ & $4.57$ & $6.37$ & $38.75$ & $0.9333$ & $3.76$ & $5.07$ & $36.41$ & $0.9101$ & $4.64$ & $6.38$ \\
HiNet\cite{jing2021hinet} & $46.46$ & $0.9633$ & $2.39$ & $3.46$ & $48.34$ & $0.9676$ & $1.98$ & $2.80$ & $46.47$ & $0.9622$ & $2.44$ & $3.47$ \\
StegFormer\cite{ke2024stegformer} & $48.91$ & $0.9889$ & $1.43$ & $2.49$ & $52.86$ & $\textbf{0.9941}$ & $0.95$ & $1.60$ & $49.14$ & $0.987$ & $1.51$ & $2.47$ \\
Ours & $\textbf{50.21}$ & $\textbf{0.9895}$ & $\textbf{1.39}$ & $\textbf{2.37}$ & $\textbf{54.62}$ & $0.9447$ & $\textbf{0.90}$ & $\textbf{1.44}$ & $\textbf{50.45}$ & $\textbf{0.9877}$ & $\textbf{1.46}$ & $\textbf{2.35}$
\\
\bottomrule
\end{tabularx}
\end{table*}

\begin{table*}[htbp]
\centering
\caption{benchmark compare of ${Secret}$/${Recovery}$ image pair on different datasets}
\label{table:comparison2}
\begin{tabularx}{\textwidth}{lXXXX|XXXX|XXXX}
\toprule
\multirow{2}{*}{} & \multicolumn{4}{c|}{COCO${(256\times256)}$} & \multicolumn{4}{c|}{DIV2K${(256\times256)}$} & \multicolumn{4}{c}{ImageNet${(256\times256)}$} \\
\cmidrule{2-13}
 & PSNR$\uparrow$ & SSIM$\uparrow$ & MAE$\downarrow$  & RMSE$\downarrow$
 & PSNR$\uparrow$ & SSIM$\uparrow$ & MAE$\downarrow$  & RMSE$\downarrow$ 
 & PSNR$\uparrow$ & SSIM$\uparrow$ & MAE$\downarrow$  & RMSE$\downarrow$ \\
\midrule
HiDDeN\cite{zhu2018hidden} & $30.51$ & $0.8392$ & $8.43$ & $11.29$ & $30.75$ & $0.8422$ & $8.31$ & $11.04$ & $30.38$ & $0.8383$ & $8.66$ & $11.66$ \\
Weng\cite{weng2019high} & $34.84$ & $0.9053$ & $4.95$ & $6.85$ & $35.53$ & $0.9155$ & $4.67$ & $6.39$ & $34.75$ & $0.902$ & $5.05$ & $6.98$ \\
HiNet\cite{jing2021hinet} & $47.72$ & $0.9806$ & $1.76$ & $2.60$ & $50.08$ & $0.9872$ & $1.51$ & $\textbf{2.26}$ & $47.77$ & $0.9798$ & $1.78$ & $2.62$ \\
StegFormer\cite{ke2024stegformer} & $49.25$ & $0.9867$ & $1.56$ & $2.46$ & $50.45$ & $0.9881$ & $1.48$ & $2.32$ & $49.27$ & $0.9865$ & $1.57$ & $2.47$ \\
Ours & $\textbf{49.63}$ & $\textbf{0.9874}$ & $\textbf{1.52}$ & $\textbf{2.39}$ & $\textbf{50.61}$ & $\textbf{0.9886}$ & $\textbf{1.45}$ & $2.27$ & $\textbf{49.62}$ & $\textbf{0.9872}$ & $\textbf{1.52}$ & $\textbf{2.39}$
\\
\bottomrule
\end{tabularx}
\end{table*}

\begin{figure*}[h]
\centering
\begin{subfigure}[b]{0.16\textwidth}
    \centering
    \caption*{Original}
    \includegraphics[width=\textwidth]{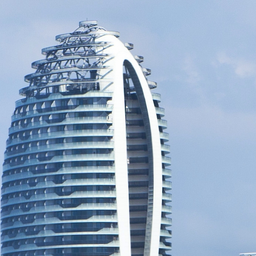}
    \caption*{(PSNR/SSIM)}
    \includegraphics[width=\textwidth]{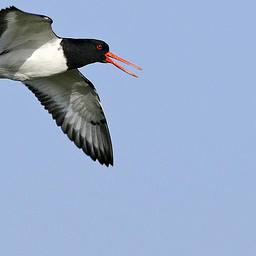}
    \caption*{(PSNR/SSIM)}
\end{subfigure}
\hfill
\begin{subfigure}[b]{0.16\textwidth}
    \centering
    \caption*{HiDDeN\cite{zhu2018hidden}}
    \includegraphics[width=\textwidth]{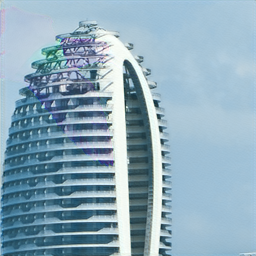}
    \caption*{(36.34/0.9732)}
    \includegraphics[width=\textwidth]{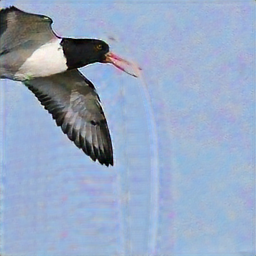}
    \caption*{(33.53/0.8329)}
\end{subfigure}
\hfill
\begin{subfigure}[b]{0.16\textwidth}
    \centering
    \caption*{Weng\cite{weng2019high}}
    \includegraphics[width=\textwidth]{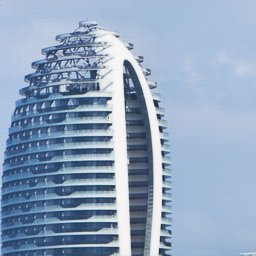}
    \caption*{(38.06/0.9526)}
    \includegraphics[width=\textwidth]{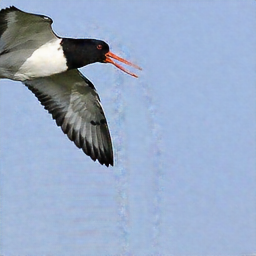}
    \caption*{(38.57/0.9247)}
\end{subfigure}
\hfill
\begin{subfigure}[b]{0.16\textwidth}
    \centering
    \caption*{HiNet\cite{jing2021hinet}}
    \includegraphics[width=\textwidth]{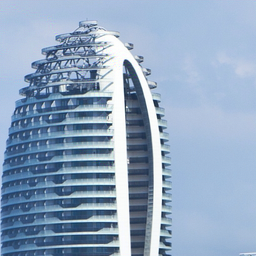}
    \caption*{(52.26/0.9890)}
    \includegraphics[width=\textwidth]{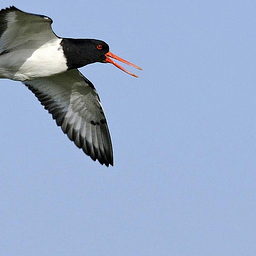}
    \caption*{(53.53/0.9927)}
\end{subfigure}
\hfill
\begin{subfigure}[b]{0.16\textwidth}
    \centering
    \caption*{StegFormer\cite{ke2024stegformer}}
    \includegraphics[width=\textwidth]{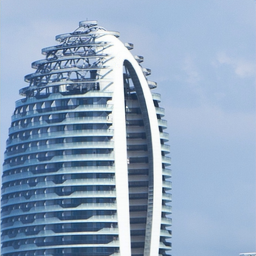}
    \caption*{(54.04/0.9976)}
    \includegraphics[width=\textwidth]{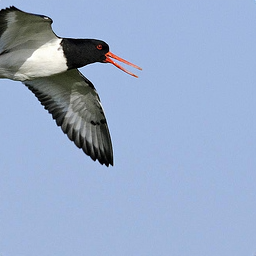}
    \caption*{(55.27/0.9956)}
\end{subfigure}
\hfill
\begin{subfigure}[b]{0.16\textwidth}
    \centering
    \caption*{Ours}
    \includegraphics[width=\textwidth]{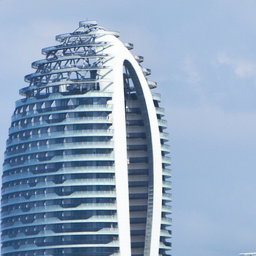}
    \caption*{(55.31/0.9980)}
    \includegraphics[width=\textwidth]{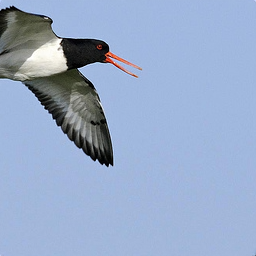}
    \caption*{(53.54/0.9957)}
\end{subfigure}
\caption{Visualization comparison of different Steganography methods on COCO. The higher the value of PSNR and SSIM, the better the model performance.}
\label{fig:comparison}
\end{figure*}

\section{Experiment}
\subsection{Dataset}
During training, we use the DIV2K dataset \cite{agustsson2017ntire} to train our model. The sizes of the datasets used are as follows: the COCO dataset contains 30,000 images, the DIV2K dataset contains 800 images, and the ImageNet dataset contains 50,000 images. For testing, we use the DIV2K, COCO \cite{lin2014microsoft}, and ImageNet \cite{deng2009imagenet} datasets to evaluate the model's generalization capabilities.
\subsection{Training Details}
We use the network architecture designed in StegFormer \cite{ke2024stegformer} as our backbone. We employed AdamW \cite{loshchilov2017decoupled} as the optimizer, initializing the learning rate at 1e-3 and gradually reducing it to 1e-6. To enhance the model's performance and generalization capabilities, we applied data augmentation to the dataset. Specifically, we performed random cropping with a crop size of 256×256, horizontal flipping, and random rotation within an angle range of $[0°,90°]$.

\begin{table*}[htbp]
 \centering
 \caption{LPIPS$\downarrow$ compare on different method}
 \label{table:comparison3}
 \begin{tabularx}{\textwidth}{lXX|XX|XX}
 \toprule
 \multirow{2}{*}{} & \multicolumn{2}{c|}{COCO2017${(256\times256)}$} & \multicolumn{2}{c|}{DIV2K${(256\times256)}$} & \multicolumn{2}{c}{ImageNet${(256\times256)}$} \\
 \cmidrule{2-7} 
  & Cover/Stego & Secret/Recovery
  & Cover/Stego & Secret/Recovery
  & Cover/Stego & Secret/Recovery \\
 \midrule
HiDDeN\cite{zhu2018hidden} & $0.1021$ & $0.1244$ & $0.0763$ & $0.1226$ & $0.1043$ & $0.1246$ \\
Weng\cite{weng2019high} & $0.0664$ & $0.0825$ & $0.0459$ & $0.0766$ & $0.0672$ & $0.0839$\\
HiNet\cite{jing2021hinet} & $0.0158$ & $0.0152$ & $0.0112$ & $\textbf{0.0106}$ & $0.0162$ & $0.0157$\\
StegFormer\cite{ke2024stegformer} & $0.0129$ & $0.0139$ & $0.0077$ & $0.0121$ & $0.0129$ & $0.0143$\\
Ours & $\textbf{0.0109}$ & $\textbf{0.0130}$ & $\textbf{0.0060}$ & $0.0114$ & $\textbf{0.0109}$ & $\textbf{0.0134}$\\
\bottomrule
\end{tabularx}
\end{table*}

\begin{figure*}[htbp]
    \centerline{
    \hspace{1.25mm}
    \begin{minipage}{0.26\textwidth}
        \includegraphics[width=\textwidth]{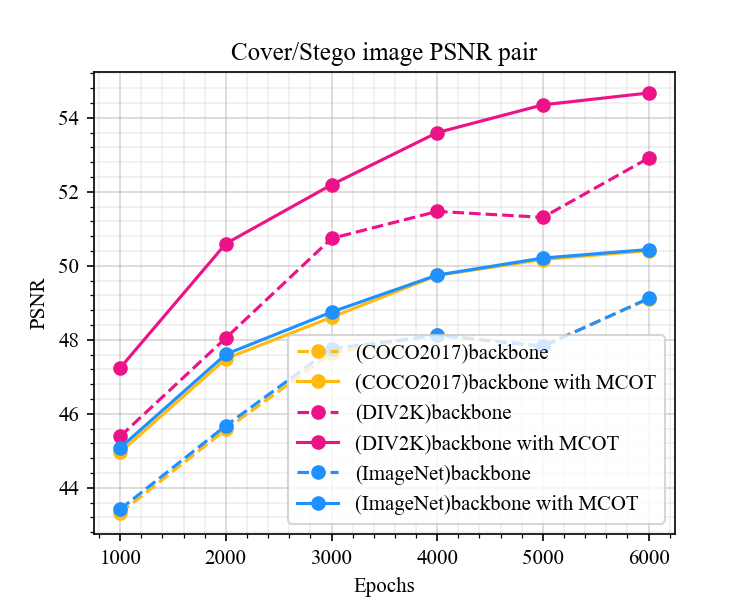}
    \end{minipage}\hfill
    \begin{minipage}{0.26\textwidth}
        \includegraphics[width=\textwidth]{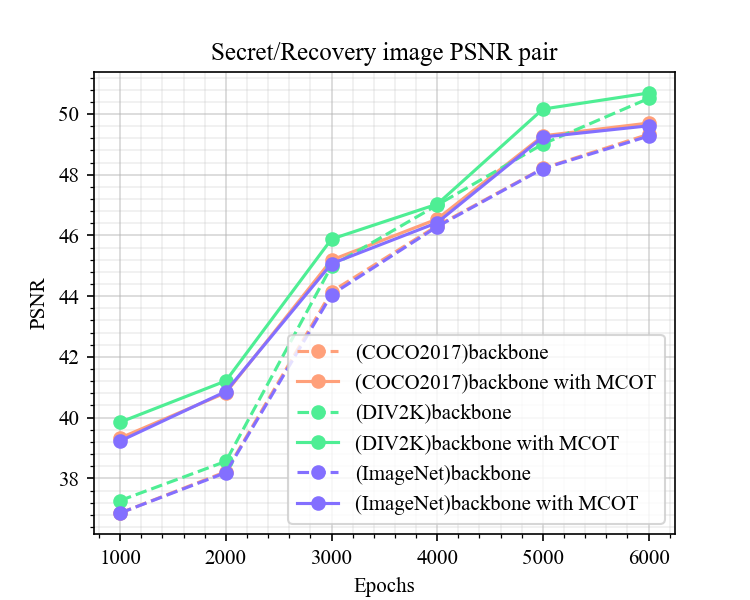}
    \end{minipage}\hfill
    \begin{minipage}{0.26\textwidth}
        \includegraphics[width=\textwidth]{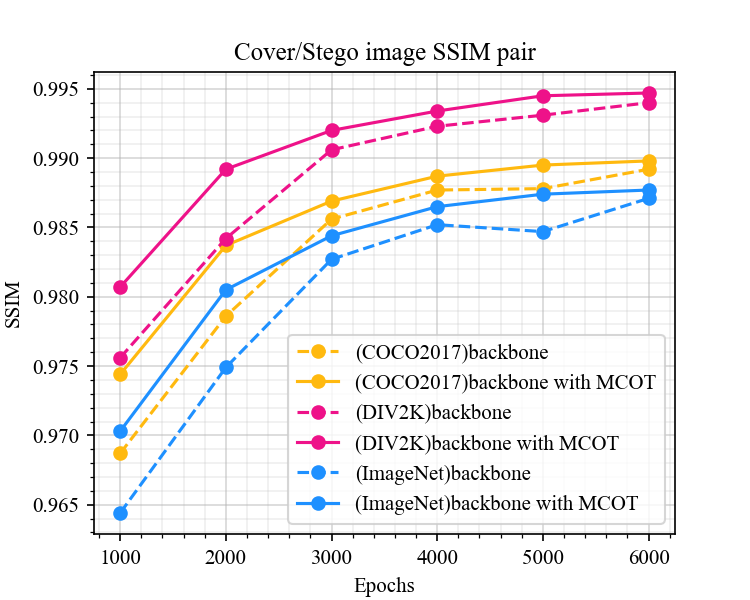}
    \end{minipage}\hfill
    \begin{minipage}{0.26\textwidth}
        \includegraphics[width=\textwidth]{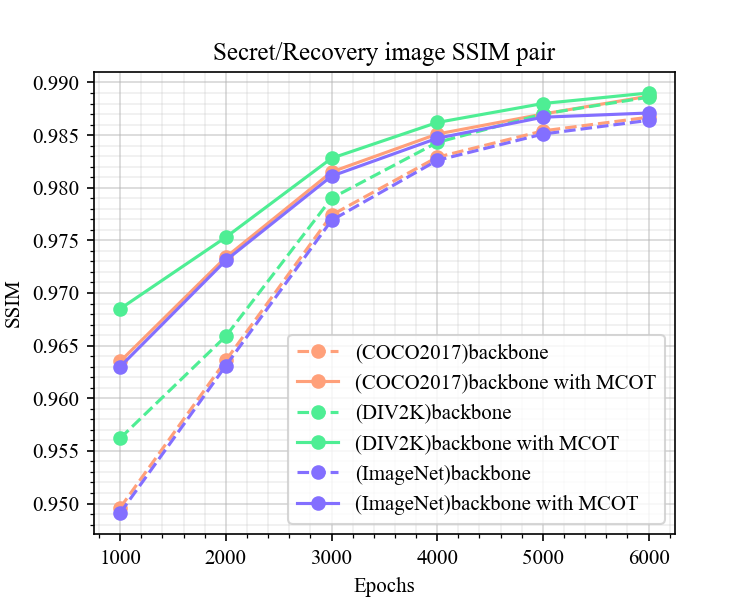}
    \end{minipage}
    }

\caption{The models utilizing MCOT (solid lines in the figure) exhibit higher PSNR values across all datasets compared to models without MCOT (dashed lines). Similarly, the SSIM values for models using MCOT are higher across all datasets than those for models not using MCOT.}
\label{fig:ablation reserach}
\end{figure*}

\subsection{Evaluation Metrics}
We primarily use PSNR, SSIM and LPIPS metrics to evaluate the performance of steganography. Note that following previous work \cite{jing2021hinet,ke2024stegformer}, we assess PSNR on the Y channel of the YCbCr color space for more accurate evaluation.

\begin{figure}[htbp]
\centerline{
\hspace{3mm}
\includegraphics[scale=0.25]{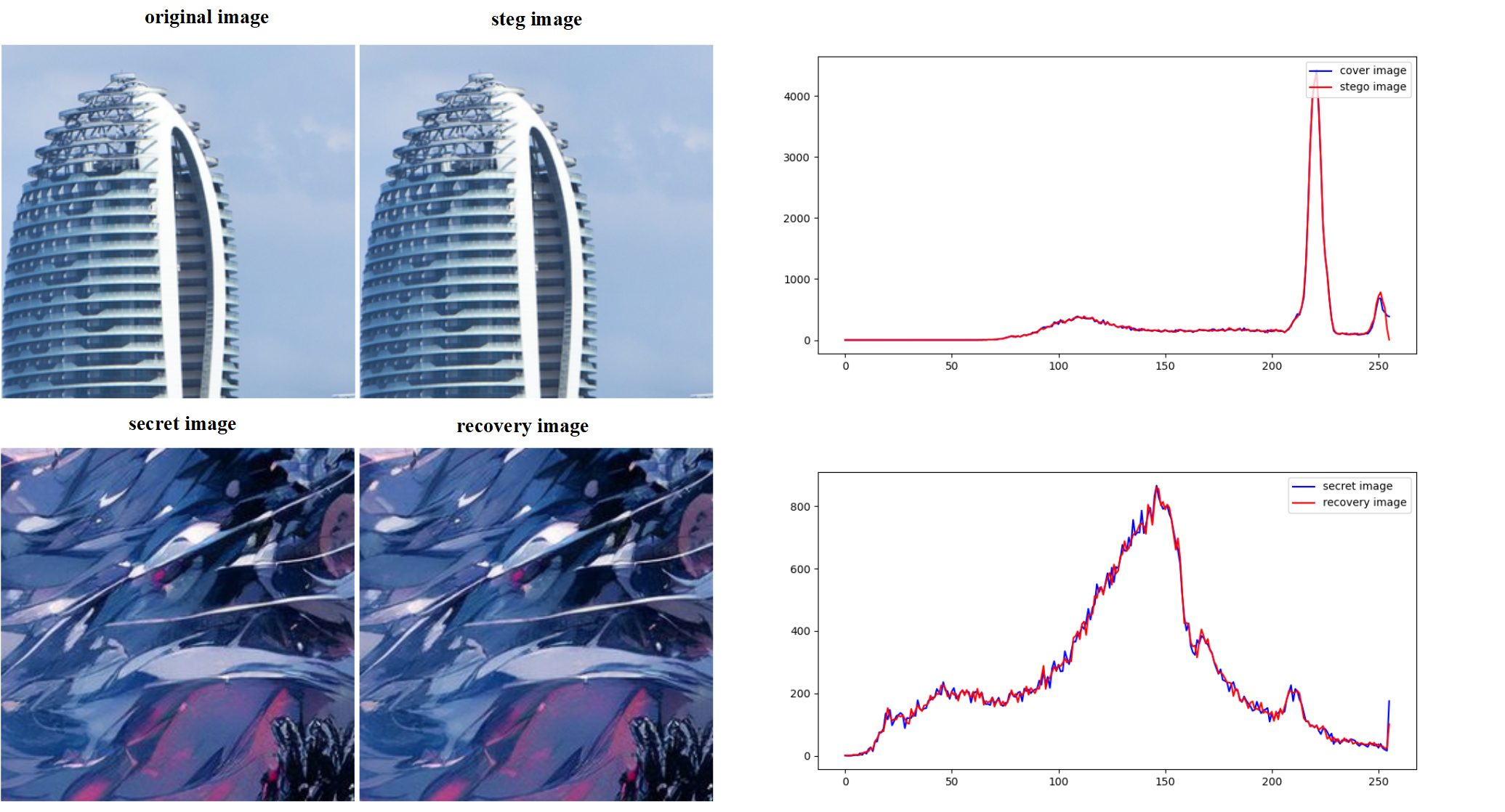}
}
\caption{The gray histogram comparison between cover/secret image and stego/recovery image.
}
\label{fig:visualize1}
\end{figure}

\begin{figure}[tb] \centering
    \includegraphics[width=0.45\textwidth,height=0.33\textwidth]{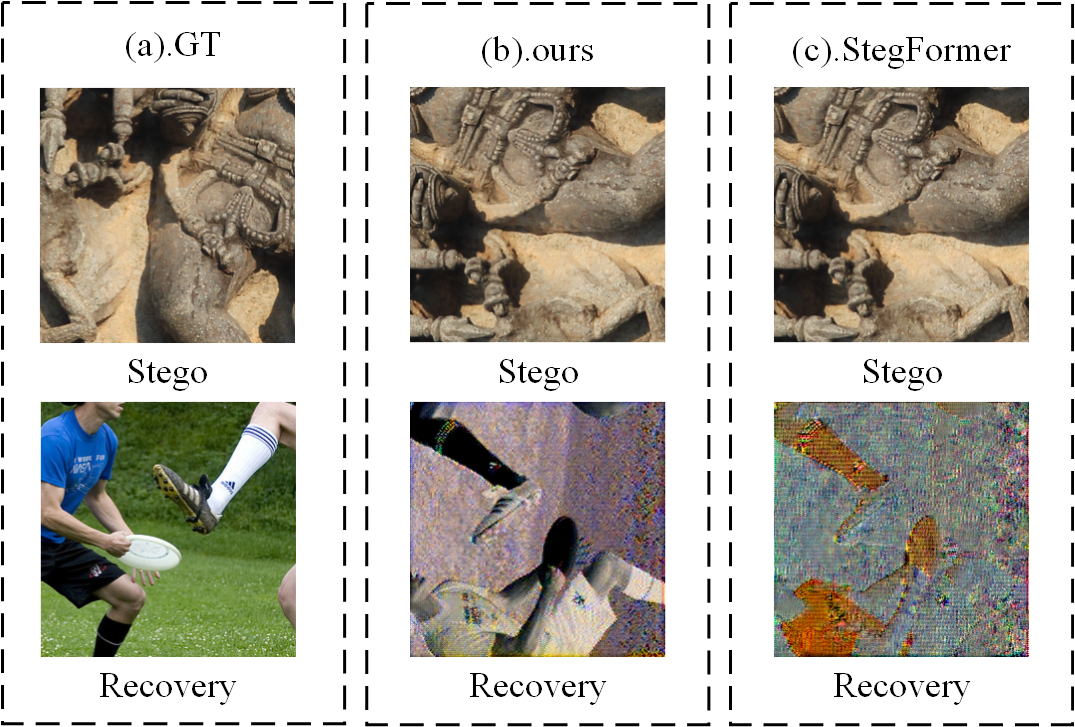}
    \caption{(a) shows the Stego image and the Recovery image under normal conditions. (b)$\&$(c) shows the extraction of Recovery images when Stego is disturbed.} \label{fig:robustness}
    \vspace{-1em}
\end{figure}

\subsection{Steganography Compare}
Table~\ref{table:comparison1} compares the numerical results of our StegOT under ideal conditions with those of current mainstream methods. Note that the pixel dimensions of the cover and secret images used in our experiments are ${256\times256}$. Specifically, for the cover/secret image pairs, our StegOT achieved a 1.3dB improvement in PSNR over StegFormer on the DIV2K, COCO, and ImageNet datasets. Similarly, as shown in Table~\ref{table:comparison2}, the PSNR performance for the secret/recovered image pairs was also enhanced on all three datasets. Furthermore, Table~\ref{table:comparison3} shows that our method achieves the lowest LPIPS value across all test datasets, with an outstanding result of 0.006 for the Cover/Stego pair on the DIV2K dataset.

\subsection{Ablation Research}
We remove MCOT from the model. As shown in Figure \ref{fig:ablation reserach}, it is evident that MCOT is crucial for StegOT. With MCOT, both the cover/stego and secret/recovery pairs show significant improvements in PSNR and SSIM. This improvement occurs because the latent vectors generated by MCOT contain more effective information, demonstrating that optimal transport effectively mitigates the impact of mode collapse. Meanwhile, We have conducted robustness experiments for both StegOT and StegFormer\cite{ke2024stegformer}, as shown in Figure~\ref{fig:robustness}. In these experiments, we applied a rotation transformation to the stego images. The second column displays the watermarks extracted from the stego images. While the performance results are not as strong as anticipated, we believe that the optimal transport approach has successfully embedded more information from the secret image into the stego image. As a result, even when the stego image is subjected to interference, the performance remains stronger compared to StegFormer.

\subsection{Visualization}
Figure~\ref{fig:comparison} demonstrates the steganography capabilities of different methods. The first column shows the cover and secret images. In the other columns, the images in the first row are the stego images generated by different methods, and the images in the second row are the recovery images. The (PSNR/SSIM) in the first row compares the cover image to the stego image, while the (PSNR/SSIM) in the second row compares the secret image to the recovery image. Figure~\ref{fig:visualize1} shows that the histograms of cover/stego and secret/recovery remain almost unchanged, indicating that the characteristics of the images are largely preserved when embedding two images.

\section{Conclusion}
In this study, we proposed a steganography model based on optimal transport theory, named MCOT, to optimize the information balance between cover and secret images. Our model uses an encoder to map images into latent space and an optimal transport module to transform a multi-peak feature distribution into a single-peak distribution, enabling the latent vector to contain both cover and secret image information. Various experiments show that the introduction of the optimal transmission theory can effectively improve the steganographic performance and robustness of the model.

\section*{Acknowledgment}
This work was in part supported by Research Capacity Enhancement Program for Young and Mid-career Teachers from Universities in Guangxi (2025KY0253), Innovation Project of Guangxi Graduate Education (Grant No. YCSW2024318), Guangxi Key Technologies R$\&$D Program (2024AB17004), Guilin Scientific Research and Technology Development Program(20230120-3), The Guangxi Science and Technology Major Project (Grant No.AA22068057) and Key Topics of Guangxi Science and Technology Think Tank (Gui Science Association(2024)K-148).

\bibliographystyle{./IEEEtran}
\bibliography{IEEEexample}

\vspace{12pt}

\end{document}